\documentclass[10pt,twocolumn,letterpaper]{article}

\usepackage[pagebackref=true,breaklinks=true,letterpaper=true,colorlinks,bookmarks=false]{hyperref}
\usepackage{cvpr}
\usepackage{times}
\usepackage{epsfig}
\usepackage{graphicx}
\usepackage{amsmath}
\usepackage{subfig}
\usepackage{amssymb}
\usepackage{tabularx}
\usepackage{booktabs}
\usepackage{multirow}
\usepackage{enumitem}
\usepackage{multicol}
\usepackage{csquotes}
\usepackage{rotating}
\usepackage{floatrow}
\usepackage{array, makecell}
\usepackage[flushleft]{threeparttable} 
\usepackage{algorithm}
\usepackage{algpseudocode}

\newcommand{\ra}[1]{\renewcommand{\arraystretch}{#1}}
\newlength{\tempdima}
\newcommand{\rowname}[1]
{\rotatebox{90}{\makebox[\tempdima][c]{\textbf{#1}}}}

\algblock{Input}{EndInput}
\algnotext{EndInput}
\algblock{Output}{EndOutput}
\algnotext{EndOutput}
\newcommand{\Desc}[2]{\State \makebox[2em][l]{#1}#2}

\cvprfinalcopy 
\newcommand{\norm}[1]{\left\lVert#1\right\rVert}
\newcommand{\IE}{\mathbb{E}}
\newcommand{\EL}{\mathcal{L}}

\def\cvprPaperID{****} 

\ifcvprfinal\fi
\begin{document}

\title{AdvFaces: Adversarial Face Synthesis}

\author{Debayan Deb\\
Michigan State University\\
East Lansing, MI, USA\\
{\tt\small debdebay@msu.edu}
\and
Jianbang Zhang\\
Lenovo Inc.\\
Morrisville, NC, USA\\
{\tt\small zhangjb2@lenovo.com}
\and
Anil K. Jain\\
Michigan State University\\
East Lansing, MI, USA\\
{\tt\small jain@cse.msu.edu}
}

\twocolumn[{%
\renewcommand\twocolumn[1][]{#1}%
\maketitle
\begin{center}
\vspace{-1.2em}
    \centering
    \captionsetup{font=small}
    \begin{minipage}{0.18\linewidth}
    \includegraphics[width=\linewidth]{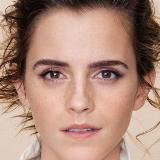}{\footnotesize \\ \centering}\\[0.1em]
    \includegraphics[width=\linewidth]{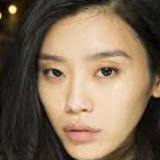}{\footnotesize \\ \centering}\\
    \centering {\small(a) Enrolled Face}
    \end{minipage}\;
    \begin{minipage}{0.18\linewidth}
    \centering\includegraphics[width=\linewidth]{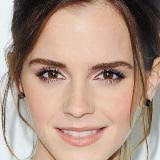}{\footnotesize \\ \centering0.72}\\[0.1em]
    \includegraphics[width=\linewidth]{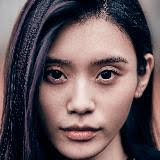}{\footnotesize \\ \centering0.78}\\
    \centering {\small(b) Input Probe}
    \end{minipage}\;
    \begin{minipage}{0.18\linewidth}
    \centering\includegraphics[width=\linewidth]{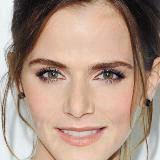}{\footnotesize \\ \centering0.22}\\[0.1em]
    \includegraphics[width=\linewidth]{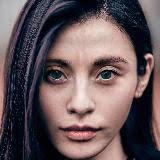}{\footnotesize \\ \centering0.12}\\
    \centering {\small(c) AdvFaces}
    \end{minipage}\;
    \begin{minipage}{0.18\linewidth}
    \centering\includegraphics[width=\linewidth]{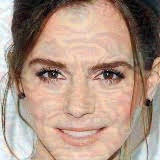}{\footnotesize \\ \centering0.26}\\[0.1em]
    \includegraphics[width=\linewidth]{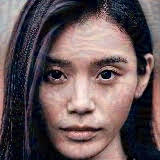}{\footnotesize \\ \centering0.25}\\
    \centering {\small(d) PGD~\cite{madry}}
    \end{minipage}\;
    \begin{minipage}{0.18\linewidth}
    \centering\includegraphics[width=\linewidth]{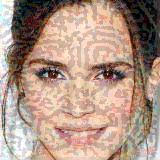}{\footnotesize \\ \centering0.14}\\[0.1em]
    \includegraphics[width=\linewidth]{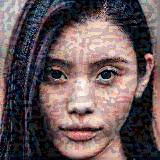}{\footnotesize \\ \centering0.25}\\
    \centering {\small(e) FGSM~\cite{goodfellow}}
    \end{minipage}
    \captionof{figure}{Example gallery and probe face images (source: \url{https://bit.ly/2LN7J50}) and corresponding synthesized adversarial examples. (a) Two celebrities' real face photo enrolled in the gallery and (b) the same subject's probe image; (c) Adversarial examples generated from (b) by our proposed synthesis method, AdvFaces; (d-e) Results from two state-of-the-art adversarial example generation methods. Cosine similarity scores ($\in[-1,1]$) obtained by comparing (b-e) to the enrolled image in the gallery via ArcFace~\cite{arcface} are shown below the images. A score above $0.28$ (threshold @ $0.1\%$ False Accept Rate) indicates that two face images belong to the same subject. Here, a successful obfuscation attack would mean that humans can identify the adversarial probes and enrolled faces as belonging to the same identity but an automated face recognition system considers them to be from different subjects. The proposed AdvFaces automatically learns to perturb those facial regions (set of pixels) that will evade an automated face recognition system, while the other baselines perturb each pixel in the image.}
    \label{fig:frontpage}
\end{center}%
}]

\begin{abstract}
   Face recognition systems have been shown to be vulnerable to adversarial examples resulting from adding small perturbations to probe images. Such adversarial images can lead state-of-the-art face recognition systems to falsely reject a genuine subject (obfuscation attack) or falsely match to an impostor (impersonation attack). Current approaches to crafting adversarial face images lack perceptual quality and take an unreasonable amount of time to generate them. We propose, AdvFaces, an automated adversarial face synthesis method that learns to generate minimal perturbations in the salient facial regions via Generative Adversarial Networks. Once AdvFaces is trained, it can automatically generate imperceptible perturbations that can evade state-of-the-art face matchers with attack success rates as high as $97.22\%$ and $24.30\%$ for obfuscation and impersonation attacks, respectively. 
\end{abstract}

\section{Introduction}
From mobile phone unlock, to boarding a flight at airports, the ubiquity of automated face recognition systems (AFR) is evident. With deep learning models, AFR systems are able to achieve accuracies as high as 99\% True Accept Rate (TAR) at 0.1\% False Accept Rate (FAR)~\cite{nist}. The model behind this success is a Convolutional Neural Network (CNN)~\cite{facenet, sphereface, arcface} and the availability of large face datasets to train the model. However, CNN models have been shown to be vulnerable to \emph{adversarial perturbations~}\footnote{Adversarial perturbations refer to altering an input image instance with small, human  imperceptible changes in a manner that can evade CNN models.}\cite{szegedy, goodfellow, universal, boosting}. Szegedy \etal first showed the dangers of \emph{adversarial examples} in the image classification domain, where perturbing the pixels in the input image can cause CNNs to misclassify the image~\cite{szegedy} even when the amount of perturbation is imperceptible to the human eye. Despite impressive recognition performance, prevailing AFR systems are still vulnerable to the growing threat of adversarial examples (see Figure~\ref{fig:frontpage}) as explained below.

\begin{figure}[!t]
    \centering
    \captionsetup{font=small}
      \begin{minipage}{\linewidth}
      \centering
        \subfloat[Print attack]{\includegraphics[width=0.32\linewidth]{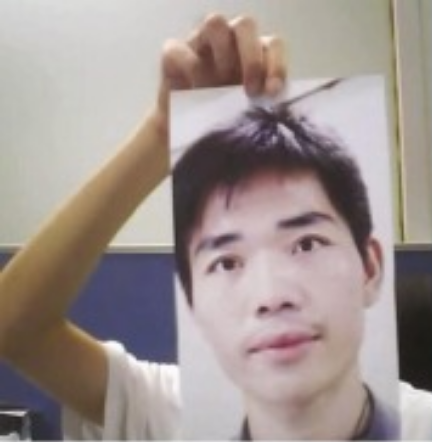}}\hfill
        \subfloat[Replay attack]{\includegraphics[width=0.32\linewidth]{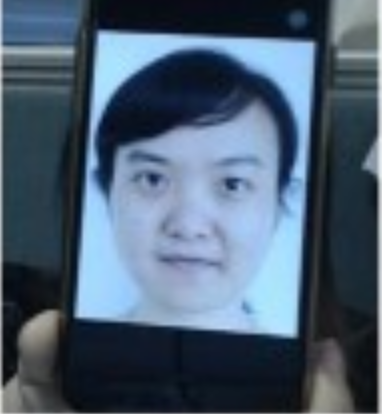}}\hfill
        \subfloat[Mask attack]{\includegraphics[width=0.32\linewidth]{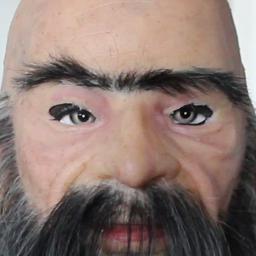}}
       \end{minipage}\\[0.5em]
       \begin{minipage}{\linewidth}
      \centering   
        \subfloat[Adversarial Faces synthesized via proposed AdvFaces]{
        \includegraphics[width=0.32\linewidth]{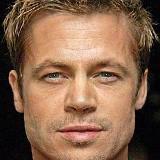}\hfill\hfill\hfill
        \includegraphics[width=0.32\linewidth]{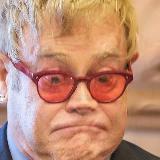}\hfill\hfill\hfill
       \includegraphics[width=0.32\linewidth]{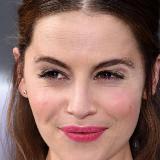}}
      \end{minipage}
      \caption{Three types of face presentation attacks: (a) printed photograph, (b) replaying the targeted person's video on a smartphone, and (c) a silicone mask of the target's face. Face presentation attacks require a physical artifact. Adversarial attacks (d), on the other hand, are digital attacks that can compromise either a probe image or the gallery itself. To a human observer, face presentation attacks (a-c) are more conspicuous than adversarial faces (d).}
    \label{fig:spoof}
\end{figure}

\begin{figure}[!t]
    \centering
    \captionsetup{font=small}
    \subfloat[Obfuscation Attack]{\includegraphics[width=\linewidth]{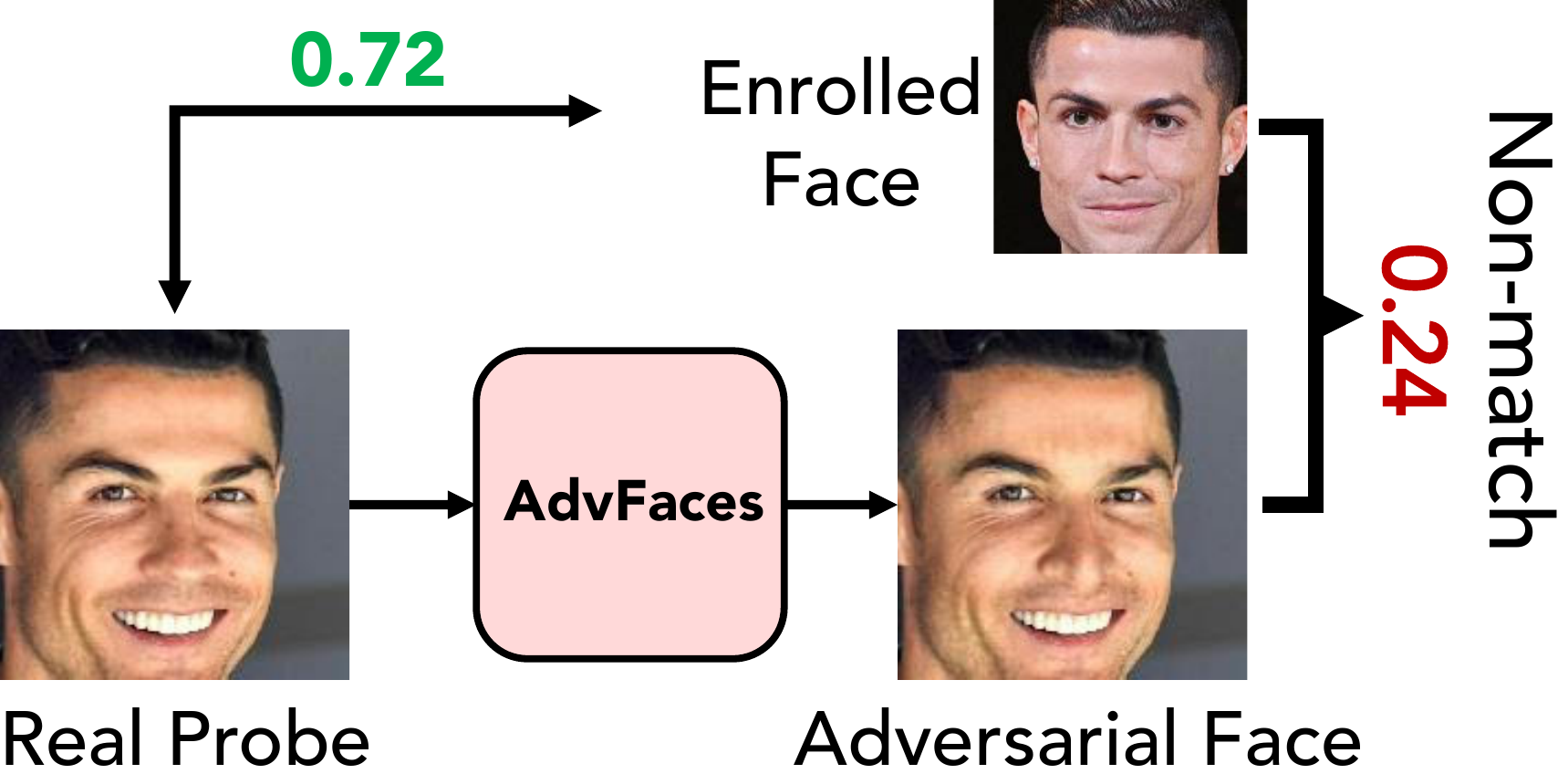}}\hfill
    \subfloat[Impersonation Attack]{\includegraphics[width=\linewidth]{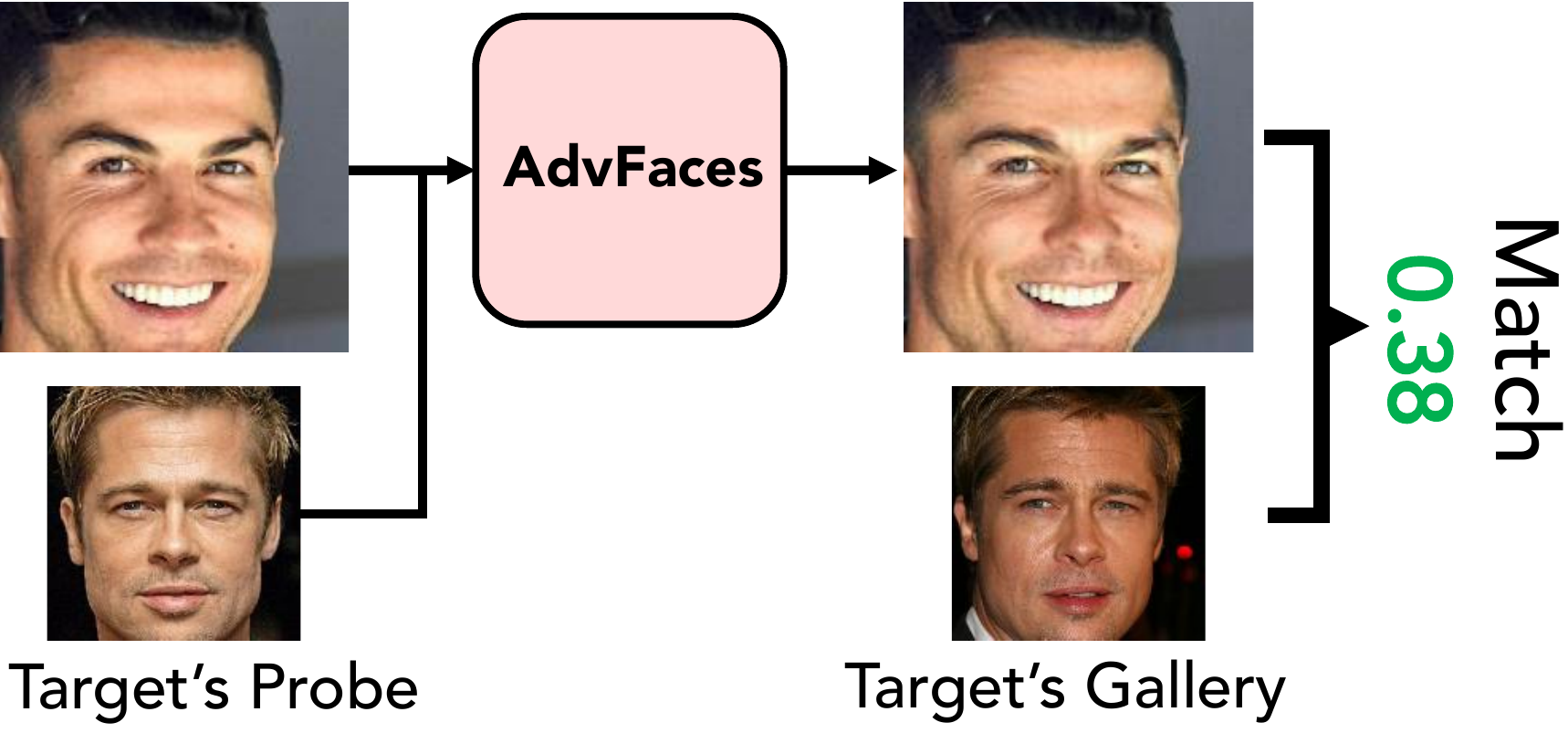}}
    \caption{Once trained, AdvFaces automatically generates an adversarial face image. During an obfuscation attack, (a) the adversarial face appears to be a benign example of Cristiano Ronaldo's face, however, it fails to match his enrolled image. AdvFaces can also combine Cristiano's probe and Brad Pitt's probe to synthesize an adversarial image that looks like Cristiano but matches Brad's gallery image (b). Cosine similarity scores obtained from ArcFace~\cite{arcface} ($0.28$ threshold @ $0.1\%$ FAR). Source:~\url{https://bit.ly/2LN7J50}}
    \label{fig:overview}
\end{figure}

A hacker can maliciously perturb his face image in a manner that can cause AFR systems to match it to a target victim (\emph{impersonation attack}) or any identity other than the hacker (\emph{obfuscation attack}). Yet to the human observer, this adversarial face image should appear as a legitimate face photo of the attacker (see Figure~\ref{fig:overview}). This is different from face~\emph{presentation  attacks}, where  the hacker assumes the  identity  of a target  by
presenting  a  fake  face  (also  known  as spoof face)  to  a  face
recognition system (see Figure~\ref{fig:spoof}). However, in the case of presentation attacks, the hacker needs to actively participate by wearing a mask or replaying a photograph/video of the genuine individual which may be conspicuous in scenarios where human operators are involved (such as airports). As discussed below, adversarial faces, do not require active participation during verification.

\begin{figure}[!t]
    \centering
    \captionsetup{font=small}
    \includegraphics[width=0.98\linewidth]{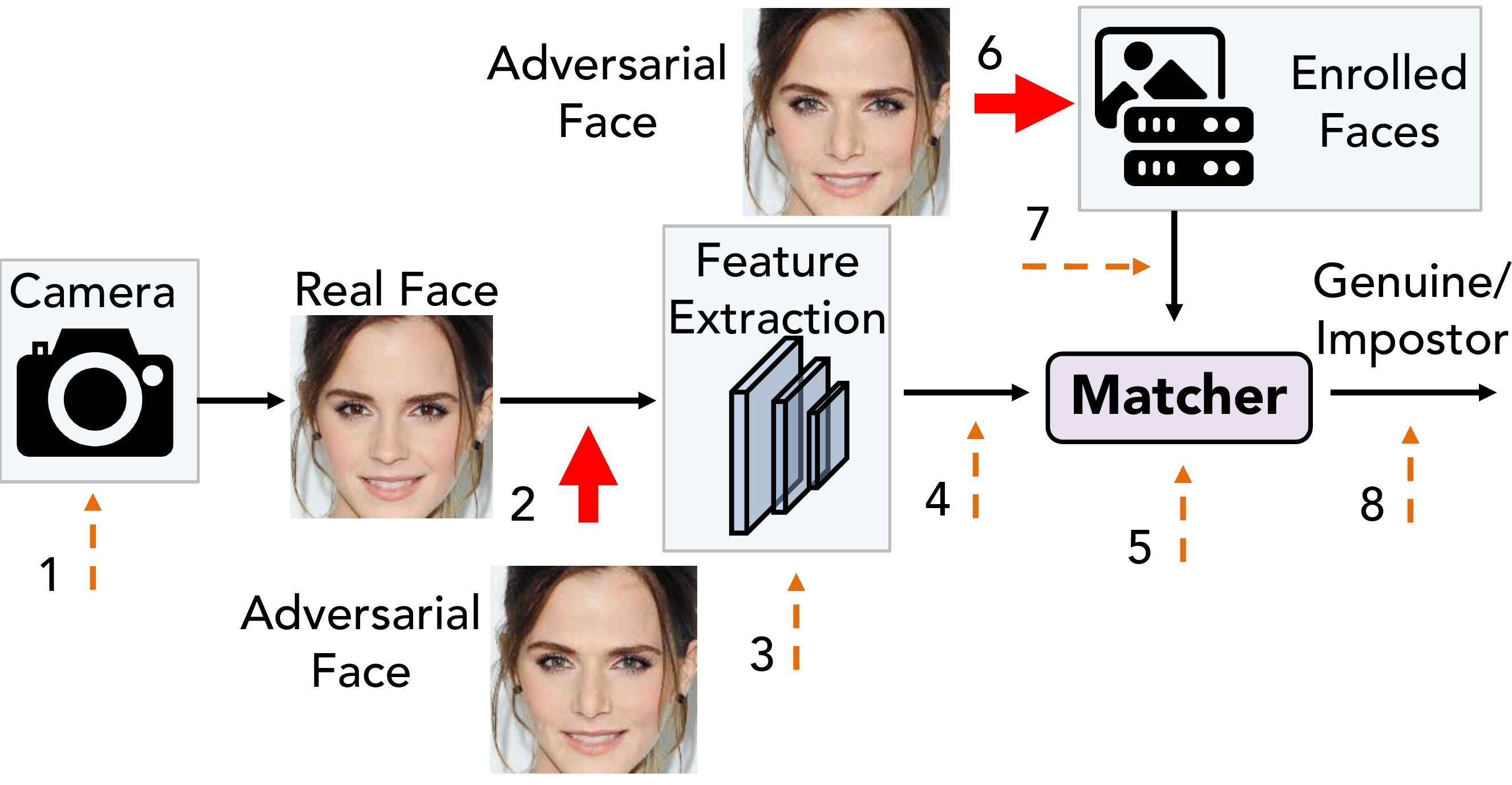}
    \caption{Eight points of attacks in an automated face recognition system~\cite{ratha}. An adversarial image can be injected in the AFR system at points 2 and 6 (solid arrows).}
    \label{fig:attacks}
\end{figure}

Consider for example, the United States Customs and Border Protection (CBP), the largest federal law enforcement agency in the United States~\cite{cbp_largest}, which (i) processes entry to the country for over a \emph{million} travellers \emph{everyday}~\cite{cbp_stats} and (ii) employs automated face recognition for verifying travelers' identities~\cite{cbp_face}. In order to evade being identified as an individual in a CBP watchlist, a terrorist can maliciously enroll an adversarial image in the gallery such that upon entering the border, his legitimate face image will be matched to a known and benign individual or to a fake identity previously enrolled in the gallery. An individual can also generate adversarial examples to dodge his own identity in order to guard personal privacy. Ratha \etal~\cite{ratha} identified eight points in a biometric system where an attack can be launched against a biometric (including face) recognition system, including AFR (see Figure~\ref{fig:attacks}). An adversarial face image can be inserted in the AFR system at point 2, where compromised face embeddings will be obtained by the feature extractor that could be used for impersonation or obfuscation attacks. The entire gallery can also be compromised if the hacker enrolls an adversarial image at point 6, where none of the probes will match to the correct identity's gallery.

Three broad categories of adversarial attacks have been identified.
\begin{enumerate}[topsep=0.5em, itemsep=-0.1em]
    \item \emph{White-box} attack: A majority of the prior work assumes full knowledge of the CNN model and then iteratively adds imperceptible perturbations to the probe image via various optimization schemes~\cite{goodfellow, madry, carlini, xiao, evtimov, papernot, kurakin, deepfool, gflm}. We posit that this is unrealistic in real-world scenarios, since the attacker may not be able to access the models.
    \item \emph{Black-box} attack: Generally, black-box attacks are launched by querying the outputs of the deployed AFR system~\cite{dong},~\cite{liu}. This may not be efficient as it may take a large number of queries to obtain a reasonable adversarial image~\cite{dong}. Further, most Commercial-Off-The-Shelf (COTS) face matchers permit only a few queries at a time to prevent such attacks.
    \item \emph{Semi-whitebox} attack: Here, a white-box model is utilized \emph{only during training} and then adversarial examples are synthesized during inference without any knowledge of the deployed AFR model. 
\end{enumerate}  
Semi-whitebox settings are appropriate for crafting adversarial faces; once the network learns to generate the perturbed instances based on a single face recognition system, attacks can be transferred to any black-box AFR systems. However, past approaches, based on Generative Adversarial Networks (GANs)~\cite{advgan, atgan, acgan}, were proposed in the image classification domain and rely on softmax probabilities~\cite{advgan, atgan, acgan, song}. Therefore, the number of object classes are assumed to be known during training and testing. In the realm of face recognition, AFR systems do not utilize the softmax layer for classification (as the number of identities are not fixed) instead features from the last fully connected layer are used for comparing face images. Song \etal proposed a GAN for generating adversarial examples specifically in the domain of face recognition, however, their method requires access to the face images enrolled in the gallery which may not be feasible in a real-world setting~\cite{song}. Other approaches for adversarial faces include adding makeup, eyeglasses, hat, or occlusions to faces~\cite{goswami, sharif_accessorize, sharif_adversarial}.

We emphasize the following requirements of the adversarial face generator:
\begin{itemize}[topsep=0.5em, itemsep=-0.1em]
    \item The generated adversarial face images should be perceptually realistic such that a human observer can identify the image as a legitimate face image pertaining to the target subject.
    \item The faces need to be perturbed in a manner such that they cannot be identified as the hacker (\emph{obfuscation attack}) or automatically matched to a target subject (\emph{impersonation attack}) by an AFR system.
    \item The amount of perturbation should be controllable by the hacker. This will allow the hacker to examine the success of the learning model as a function of amount of perturbation.
    \item The adversarial examples should be \emph{transferable} and \emph{model-agnostic} (\ie treat the target AFR model as a black-box). In other words, the generated adversarial examples should have high attack success rate on other black-box AFR systems as well.
\end{itemize}

We  propose an automated adversarial face synthesis method, named \emph{AdvFaces}, which generates an \emph{adversarial mask} for a probe face image and satisfies all the above requirements. The adversarial mask can then be added to the probe to obtain an adversarial face example that can be used either for impersonating any target identity or obfuscating one's own identity (see Figure~\ref{fig:overview}). The contributions of the paper can be summarized as follows:
\begin{enumerate}[topsep=0.5em, itemsep=-0.5em]
    \item A GAN, AdvFaces, that learns to generate visually realistic adversarial face images that are misclassified by state-of-the-art AFR systems.
    \item Adversarial faces generated via AdvFaces are model-agnostic and transferable, and achieve high success rate on 5 state-of-the-art automated face recognition systems. 
    \item Visualizing the facial regions, where pixels are perturbed and analyzing the effect of varying image resolution on the amount of perturbation.
    \item An open-source\footnote{[link omitted for blind review]} automated adversarial face generator permitting users to control the amount of perturbation.
\end{enumerate}

\section{Related Work}
\subsection{Generative Adversarial Networks (GANs)} Generative Adversarial Networks~\cite{goodfellow_gan} have success in a wide variety of image synthesis applications~\cite{dcgan, laplacian} such as style transfer~\cite{style_transfer1, style_transfer2, style_transfer3}, image-to-image translation~\cite{pix_to_pix, cycle_gan}, and representation learning~\cite{rep1, rep2, rep3}. Isola \etal showed that an image-to-image conditional GAN can vastly improve the synthesis results~\cite{pix_to_pix}. In our work, we adopt a similar adversarial loss and image-to-image network architecture in order to learn a mapping from the input face image to a perturbed output image such that the perturbed image cannot be distinguished from real face images. However, different from prior work on GANs, our objective is to synthesize face images that are not only visually realistic but are also able to evade AFR systems.

\subsection{Adversarial Attacks on Image Classification}
Majority of the published papers have focused on white-box attacks, where the hacker has full access to the target classification model that is being attacked~\cite{szegedy, goodfellow, carlini, xiao, madry}. Given an image, Goodfellow \etal proposed the Fast Gradient Sign Method (FGSM) which generates an adversarial example by back-propagating through the target model~\cite{goodfellow}. Madry \etal proposed Projected Gradient Descent (PGD), a multi-step variant of FGSM~\cite{madry}. Other works focused on optimizing adversarial perturbation by minimizing an objective function for targeted attacks while satisfying certain constraints~\cite{carlini}. However, these white-box approaches are not feasible in the face recognition domain, as the attacker may not have any knowledge of the deployed AFR system. In addition, the optimization process can require multiple queries to the target system until convergence. Instead, we propose a feed-forward network that can automatically generate an adversarial image with a single forward pass without the need for any knowledge of AFR system during inference.

Indeed, feed-forward networks have been used for synthesizing adversarial attacks. Baluja and Fischer proposed a deep autoencoder that learns to transform an input image to an adversarial image~\cite{baluja}. In their work, an $L_2$ norm loss is employed in order to constrain the generated adversarial instance to be close to the original image in the $L_2$ pixel space. In contrast, we apply a deep neural network as a discriminator that distinguishes between real and synthesized face images in order to maintain the perceptual quality of the generated adversarial examples. Studies on synthesizing adversarial instances via GANs are limited in literature~\cite{advgan, atgan, acgan}. These methods require softmax probabilities in order to evade an image classifier. However, AFR systems do not employ a softmax classification layer as the number of classes (identities) is not fixed. Instead, we propose an identity loss function better suited for generating adversarial faces using the face embeddings obtained from a face matcher.

\subsection{Adversarial Attacks on Face Recognition}
In literature, studies on generating adversarial examples in the face recognition domain are relatively limited. Bose \etal craft adversarial examples by solving constrained optimization such that a face detector cannot detect a face~\cite{bose}. In~\cite{sharif_accessorize}, perturbations are constrained to the eyeglass region of the face and adversarial image is generated by gradient-based methods. The adversarial eyeglasses can also be synthesized via generative networks~\cite{sharif_adversarial}. However, these methods rely on white-box manipulations of face recognition models, which is impractical in real-world scenarios. Dong \etal proposed an evolutionary optimization method for generating adversarial faces in black-box settings~\cite{dong}. However, they require at least 1,000 queries to the target AFR system before a realistic adversarial face can be synthesized. Song \etal employed a conditional variation autoencoder GAN for crafting adversarial face images in a semi-whitebox setting~\cite{song}. However, they only focused on impersonation attacks and require at least 5 images of the target subject for training and inference. In contrast, we train a GAN that can perform both obfuscation and impersonation attacks and requires a single face image of the target subject. 

\section{AdvFaces}
\begin{figure}[!t]
    \centering
    \captionsetup{font=small}
    \includegraphics[width=\linewidth]{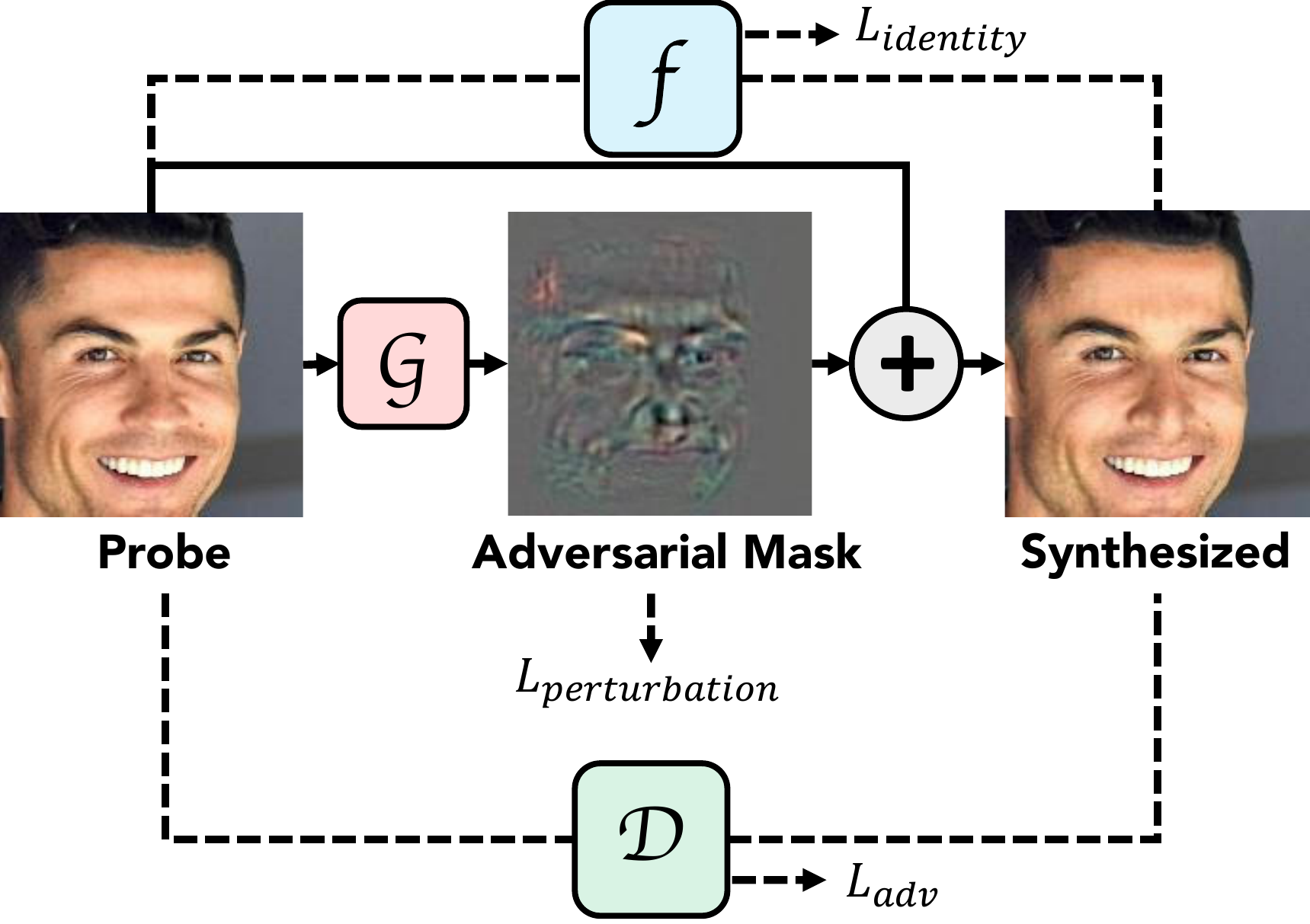}
    \caption{Overview of the proposed adversarial generation method in an obfuscation setting. Given a probe face image, AdvFaces automatically generates an adversarial mask that is then added to the probe to obtain an adversarial face image.}
    \label{fig:arch}
\end{figure}

Our goal is to synthesize a face image that visually appears to pertain to the target person, yet automatic face recognition systems either incorrectly matches the synthesized image to another person or does not match to genuine person's gallery images. Figure~\ref{fig:arch} outlines the proposed framework.

\setlength{\abovedisplayskip}{5pt}
\setlength{\belowdisplayskip}{5pt}
\setlength{\abovedisplayshortskip}{0pt}
\setlength{\belowdisplayshortskip}{0pt}

AdvFaces comprises of a generator $\mathcal{G}$, a discriminator $\mathcal{D}$, and face matcher $\mathcal{F}$ (see Figure~\ref{fig:arch} and Algorithm~\ref{alg:advfaces}).

\paragraph{Generator} The proposed generator takes an input face image, $x \in \mathcal{X}$, and outputs an image, $\mathcal{G}(x)$. The generator is conditioned on the input image $x$; for different input faces, we will get different synthesized images.

Since our goal is to obtain an adversarial image that is metrically close to the original input image, $x$, we do not wish to perturb all the pixels in the original image. For this reason, we treat the output from the generator as an additive mask and the adversarial face is $x + \mathcal{G}(x)$. If the magnitude of the pixels in $\mathcal{G}(x)$ is minimal, then the adversarial image comprises mostly of the probe $x$. Here, we denote $\mathcal{G}(x)$ as an ``adversarial mask". In order to bound the magnitude of the adversarial mask, we introduce a \emph{perturbation hinge loss} during training by minimizing the $L_2$ norm\footnote{For brevity, we denote $\IE_x \equiv \IE_{x \in \mathcal{X}}$.}:
\begin{align}
    \EL_{perturbation} &= \IE_{x}\left[\text{max}\left(\epsilon,{\norm{\mathcal{G}(x)}_2}\right)\right]
\end{align}
where $\epsilon\in [0,\infty)$ is a hyperparameter that controls the minimum amount of perturbation allowed. 

In order to achieve our goal of impersonating a target subject or obfuscating one's own identity, we need a face matcher, $\mathcal{F}$, to supervise the training of AdvFaces. For obfuscation attack, the goal is to generate an adversarial image that does not match any of the subject's gallery images. At each training iteration, AdvFaces tries to minimize the cosine similarity between face embeddings of the input probe $x$ and the generated image $x + \mathcal{G}(x)$ via an \emph{identity} loss function:
\begin{align}
    \EL_{identity} &= \IE_{x}{\left[\mathcal{F}(x, x+\mathcal{G}(x))\right]}
\end{align}
For an impersonation attack, AdvFaces maximizes the cosine similarity between the face embeddings of a randomly chosen target's probe, $y$, and the generated adversarial face $x+\mathcal{G}(x)$ via: 
\begin{align}
    \EL_{identity} &= \IE_{x}{\left[1- \mathcal{F}(y, x+\mathcal{G}(x))\right]}
\end{align}

The perturbation and identity loss functions enforce the network to learn the salient facial regions that can be perturbed minimally in order to evade automatic face recognition systems.

\paragraph{Discriminator}
Akin to previous works on GANs~\cite{goodfellow_gan, pix_to_pix}, we introduce a discriminator in order to encourage perceptual realism of the generated images. We use a fully-convolution network as a patch-based discriminator~\cite{pix_to_pix}. Here, the discriminator, $\mathcal{D}$, aims to distinguish between a probe, $x$, and a generated adversarial face image $x+\mathcal{G}(x)$ via a GAN loss:
\begin{align}
    \begin{split}
    \EL_{GAN} = &\quad\IE_{x}\left[{\log\mathcal{D}(x)}\right] +\\
    &\quad\IE_{x}{\left[\log(1-\mathcal{D}(x+\mathcal{G}(x)))\right]}
    \end{split}
\end{align}

Finally, AdvFaces is trained in an end-to-end fashion with the following objective:
\begin{align}
 \EL &= \EL_{GAN} + \lambda_{i} \EL_{identity} + \lambda_{p}\EL_{perturbation}
\end{align}
where $\lambda_i$ and $\lambda_p$ are hyper-parameters controlling the relative importance of identity and perturbation losses, respectively. Note that $\EL_{GAN}$ and $\EL_{perturbation}$ encourages the generated images to be visually similar to the original face images, while $\EL_{identity}$ optimizes for a high attack success rate.  
We train AdvFaces as a minmax game via $\{\text{arg}~\text{min}_{\mathcal{G}}\text{max}_{\mathcal{D}}\EL\}$.
After training, the generator $\mathcal{G}$ can generate an adversarial face image for any input image and can be tested on any black-box face recognition system.

\begin{table*}
\centering
\ra{1.0}
\captionsetup{font=small}
\begin{tabular}{@{}lcccccccc@{}}
\toprule \textbf{Obfuscation Attack} & \multicolumn{1}{c}{AdvFaces} & \phantom{abc}& \multicolumn{1}{c}{GFLM~\cite{gflm}} & 
\phantom{abc}& \multicolumn{1}{c}{PGD~\cite{madry}} & 
\phantom{abc} &\multicolumn{1}{c}{FGSM~\cite{goodfellow}}\\
\midrule
Attack Success Rate (\%)\\
\quad FaceNet~\cite{facenet} & 99.67 && 23.34 && 99.70 && \textbf{99.96}\\
\quad SphereFace~\cite{sphereface} & 97.22 && 29.49 && \textbf{99.34} && 98.71\\
\quad ArcFace~\cite{arcface} & \textbf{64.53} && 03.43 && 33.25 && 35.30\\
\quad COTS-A & \textbf{82.98} && 08.89 && 18.74 &&  32.48\\
\quad COTS-B & \textbf{60.71} && 05.05 && 01.49 && 18.75\\
\midrule
Structural Similarity & \textbf{0.95 $\pm$ 0.01} && 0.82 $\pm$ 0.12 && 0.29 $\pm$ 0.06 &&  0.25 $\pm$ 0.06\\
\midrule
Computation Time (s) & \textbf{0.01} && 3.22  && 11.74 && 0.03\\
\bottomrule\\
\toprule \textbf{Impersonation Attack} & \multicolumn{1}{c}{AdvFaces} &
\phantom{abc} & \multicolumn{1}{c}{A$^3$GN~\cite{song}} &\phantom{abc} & \multicolumn{1}{c}{PGD~\cite{madry}} & \phantom{abc} &\multicolumn{1}{c}{FGSM~\cite{goodfellow}}\\
\midrule
Attack Success Rate (\%)\\
\quad FaceNet~\cite{facenet} & 20.85 $\pm$ 0.40 && 05.99 $\pm$ 0.19 && \textbf{76.79 $\pm$ 0.26}  && 13.04 $\pm$ 0.12\\
\quad SphereFace~\cite{sphereface} & \textbf{20.19 $\pm$ 0.27} && 07.94 $\pm$ 0.19 && 09.03 $\pm$ 0.39 && 02.34 $\pm$ 0.03\\
\quad ArcFace~\cite{arcface} & \textbf{24.30 $\pm$ 0.44} && 17.14 $\pm$ 0.29 && 19.50 $\pm$ 1.95 && 08.34 $\pm$ 0.21\\
\quad COTS-A & \textbf{20.75 $\pm$ 0.35} && 15.01 $\pm$ 0.30 && 01.76 $\pm$ 0.10 && 01.40 $\pm$ 0.08\\
\quad COTS-B & \textbf{19.85 $\pm$ 0.28} && 10.23 $\pm$ 0.50 && 12.49 $\pm$ 0.24  && 04.67 $\pm$ 0.16 \\
\midrule
Structural Similarity & \textbf{0.92 $\pm$ 0.02} && 0.69 $\pm$ 0.04 && 0.77 $\pm$ 0.04 && 0.48 $\pm$ 0.75\\
\midrule
Computation Time (s) & \textbf{0.01} && 0.04 && 11.74 && 0.03\\
\bottomrule
\end{tabular}
\caption{Attack success rates and structural similarities between probe and gallery images for obfuscation and impersonation attacks. Attack rates for obfuscation comprises of 484,514 comparisons and the mean and standard deviation across 10-folds for impersonation reported. The mean and standard deviation of the structural similarities between adversarial and probe images along with the time taken to generate a single adversarial image (on a Quadro M6000 GPU) also reported.}
\label{tab:recognition}
\end{table*}

\begin{figure*}[!t]
\captionsetup{font=small}
\small
\settoheight{\tempdima}{\includegraphics[width=.32\textwidth]{example-image-a}}%
\centering\begin{tabular}{c@{ }c@{ }c@{ }c@{ }c@{ }c@{}}
\textbf{Gallery} & \textbf{Probe} & \textbf{AdvFaces} & \textbf{GFLM~\cite{gflm}} &\textbf{PGD~\cite{madry}} & \textbf{FGSM~\cite{goodfellow}} \\
\includegraphics[width=.15\textwidth]{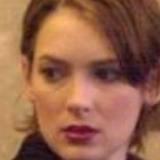}&
\includegraphics[width=.15\textwidth]{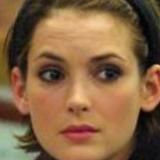}&
\includegraphics[width=.15\textwidth]{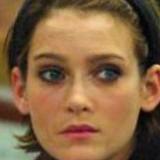}&
\includegraphics[width=.15\textwidth]{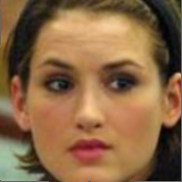}&\includegraphics[width=.15\textwidth]{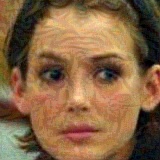} &
\includegraphics[width=.15\textwidth]{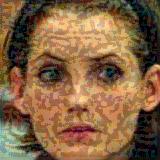}\\
 & 0.68 & 0.14 & 0.26 & 0.27 & 0.04\\
\includegraphics[width=.15\textwidth]{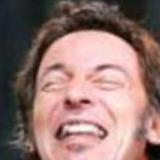}&
\includegraphics[width=.15\textwidth]{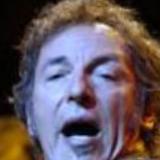}&
\includegraphics[width=.15\textwidth]{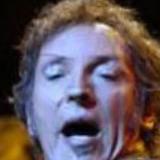}&
\includegraphics[width=.15\textwidth]{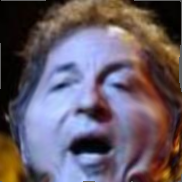}& \includegraphics[width=.15\textwidth]{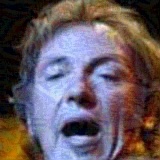} &
\includegraphics[width=.15\textwidth]{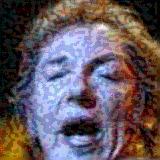}\\
 & 0.38 & 0.08 & 0.12 & 0.21 & 0.02\\
 \end{tabular}\\[0.5em]
(a) Obfuscation Attack\\[1em]
\centering\begin{tabular}{c@{ }c@{ }c@{ }c@{ }c@{ }c@{}}
\textbf{Target's Gallery} & \textbf{Target's Probe} & \textbf{Probe} & \textbf{AdvFaces} &\textbf{A$^3$GN~\cite{song}} & \textbf{FGSM~\cite{goodfellow}} \\
\includegraphics[width=.15\textwidth]{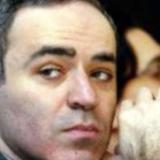}&
\includegraphics[width=.15\textwidth]{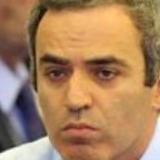}&
\includegraphics[width=.15\textwidth]{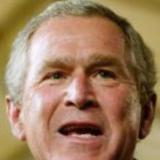} &
\includegraphics[width=.15\textwidth]{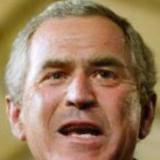}&
\includegraphics[width=.15\textwidth]{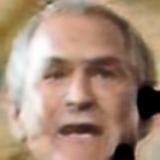}&
\includegraphics[width=.15\textwidth]{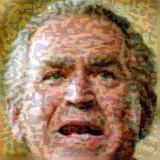}\\
 & 0.78 & 0.10 & 0.30 & 0.29 & 0.36\\
\includegraphics[width=.15\textwidth]{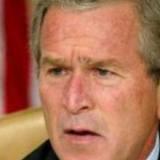}&
\includegraphics[width=.15\textwidth]{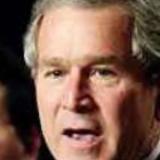}&
\includegraphics[width=.15\textwidth]{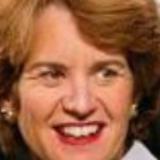}&
\includegraphics[width=.15\textwidth]{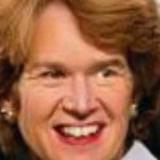}& \includegraphics[width=.15\textwidth]{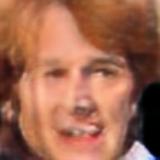}&
\includegraphics[width=.15\textwidth]{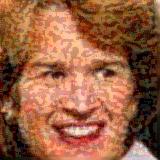}\\
 & 0.80 & 0.15 & 0.34 & 0.33 & 0.42\\
 \end{tabular}\\[0.5em]
 (b) Impersonation Attack%
\caption{Adversarial face synthesis results on LFW dataset in (a) obfuscation and (b) impersonation attack settings (cosine similarity scores obtained from ArcFace~\cite{arcface} with threshold @ $0.1\%$ FAR$=0.28$). The proposed method synthesizes adversarial faces that are seemingly inconspicuous and maintain high perceptual quality.}
\label{fig:comparison}
\end{figure*}

\section{Experimental Results}
\paragraph{Evaluation Metrics}
We quantify the effectiveness of the adversarial attacks generated by AdvFaces and other state-of-the-art baselines via (i) \emph{attack success rate} and (ii) \emph{structural similarity (SSIM)}.

The attack success rate for \emph{obfuscation attack} is computed as,
\begin{align}
   \text{Attack Success Rate} &= \frac{\text{(No. of Comparisons $< \tau$)}}{\text{Total No. of Comparisons}}
   \label{eq:attack1}
\end{align}
where each comparison consists of a subject's adversarial probe and an enrollment image. Here, $\tau$ is a pre-determined threshold computed at, say, 0.1\% FAR\footnote{We compute the threshold at $0.1\%$ FAR on all possible image pairs in LFW. For~\eg, threshold @ $0.1\%$ FAR for ArcFace is $0.28$.}. Attack success rate for \emph{impersonation attack} is defined as,
\begin{align}
   \text{Attack Success Rate} &= \frac{\text{(No. of Comparisons $\geq \tau$)}}{\text{Total No. of Comparisons}}
   \label{eq:attack2}
\end{align}
Here, a comparison comprises of an adversarial image synthesized with a target's probe and matched to the target's enrolled image. We evaluate the success rate for the impersonation setting via 10-fold cross-validation where each fold consists of a randomly chosen target.

Similar to prior studies~\cite{song}, in order to measure the similarity between the adversarial example and the input face, we compute the structural similarity index (SSIM) between the images. SSIM is a normalized metric between $-1$ (completely different image pairs) to $1$ (identical image pairs):
\begin{align}
    SSIM(x, y)\footnotemark = \frac{(2\mu_x\mu_y + c_1)(2\sigma_{xy}+c_2)}{(\mu_x^{2} + \mu^{2}_y + c_1)(\sigma_x^2 + \sigma_y^2+c_2)}
    \label{eq:ssim}
\end{align}
\footnotetext{Here, $x$ and $y$ are the two images that are compared, $\mu_x$, $\mu_y$, $\sigma^2_x$, $\sigma^2_y$, are the means and variances of $x$ and $y$, respectively. The covariance of $x$ and $y$ is $\sigma_{xy}$. Parameters $c_i = (k_iL)^2$, where $L = \left(2^{(\texttt{\# of bits per pixel})} - 1\right)$, $k_1 = 0.01$, $k_2 = 0.03$ by default~\cite{ssim}.}
\vspace{-1em}\paragraph{Datasets} We train AdvFaces on CASIA-WebFace~\cite{casia} and then test on LFW~\cite{lfw}\footnote{Training on CASIA-WebFace and evaluating on LFW is a common approach in face recognition literature~\cite{arcface, sphereface}}.
\begin{itemize}[topsep=0em, itemsep=-0.1em]
    \item \textbf{CASIA-WebFace}~\cite{casia} comprises of 494,414 face images belonging to 10,575 different subjects.
    \item \textbf{LFW}~\cite{lfw} contains 13,233 web-collected images of 5,749 different subjects. In order to compute the attack success rate, we only consider subjects with at least two face images. After this filtering, 9,614 face images of 1,680 subjects are available for evaluation.
\end{itemize}

\begin{figure}[!t]
\setlength\tabcolsep{0px}
\newcolumntype{Y}{>{\centering\arraybackslash}X}
    \centering
    \captionsetup{font=small}
    \begin{tabularx}{\linewidth}{YYYYYY}
    \toprule
        Input & w/o $\EL_{GAN}$  & w/o $\EL_{prt}$ & w/o $\EL_{idt}$ & with all  \\
    \midrule
        \includegraphics[width=0.97\linewidth]{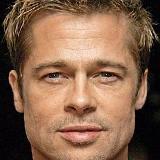} & 
        \includegraphics[width=0.97\linewidth]{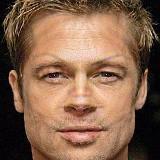} &
        \includegraphics[width=0.97\linewidth]{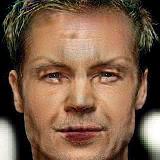} &
        \includegraphics[width=0.97\linewidth]{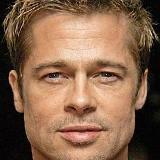} &
        \includegraphics[width=0.97\linewidth]{fig/1_g.jpg} \\
        \includegraphics[width=0.97\linewidth]{fig/3_o.jpg} & 
        \includegraphics[width=0.97\linewidth]{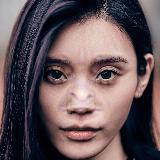} &
        \includegraphics[width=0.97\linewidth]{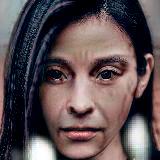} &
        \includegraphics[width=0.97\linewidth]{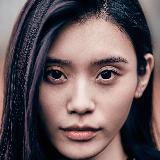} &
        \includegraphics[width=0.97\linewidth]{fig/3_g.jpg} \\
    \bottomrule
    \end{tabularx}
    \caption{Variants of AdvFaces trained without GAN loss, perturbation loss, and identity loss, respectively.}
    \label{fig:ablation}
\end{figure}

\begin{figure}[!t]
\captionsetup{font=small}
\small
\settoheight{\tempdima}{\includegraphics[width=.24\textwidth]{example-image-a}}%
\centering\begin{tabular}{@{}c@{ }c@{ }c@{ }c@{ }c@{}}
\textbf{Probe} & \textbf{Adv. Mask} & \textbf{Visualization} & \textbf{Adv. Image} \\
\includegraphics[width=.23\textwidth]{fig/3_o.jpg}&
\includegraphics[width=.23\textwidth]{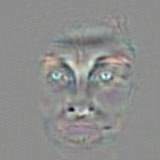}&
\includegraphics[width=.23\textwidth]{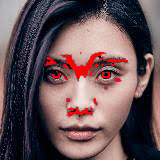}&
\includegraphics[width=.23\textwidth]{fig/3_g.jpg}\\
& & &\footnotesize 0.12\\
\includegraphics[width=.23\textwidth]{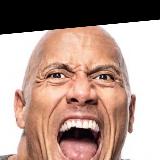}&
\includegraphics[width=.23\textwidth]{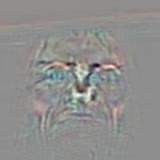}&
\includegraphics[width=.23\textwidth]{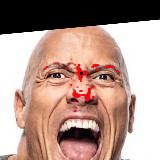}&
\includegraphics[width=.23\textwidth]{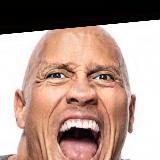}\\
& & & {\footnotesize 0.26}%
\end{tabular}
\caption{Pixels that have been perturbed (Column 3) to generate the corresponding adversarial images for the given probes (Column 1). AdvFaces outputs adversarial masks (Column 2) which are added to the probes to obtain adversarial images in the last column. State-of-the-art face matchers can be evaded by slightly perturbing salient facial regions, such as eyebrows, eyeballs, and nose (cosine similarity obtained via ArcFace~\cite{arcface}).}%
\label{fig:visualization}
\end{figure}

\begin{figure}[!t]
    \centering
    \captionsetup{font=small}
    \includegraphics[width=\linewidth]{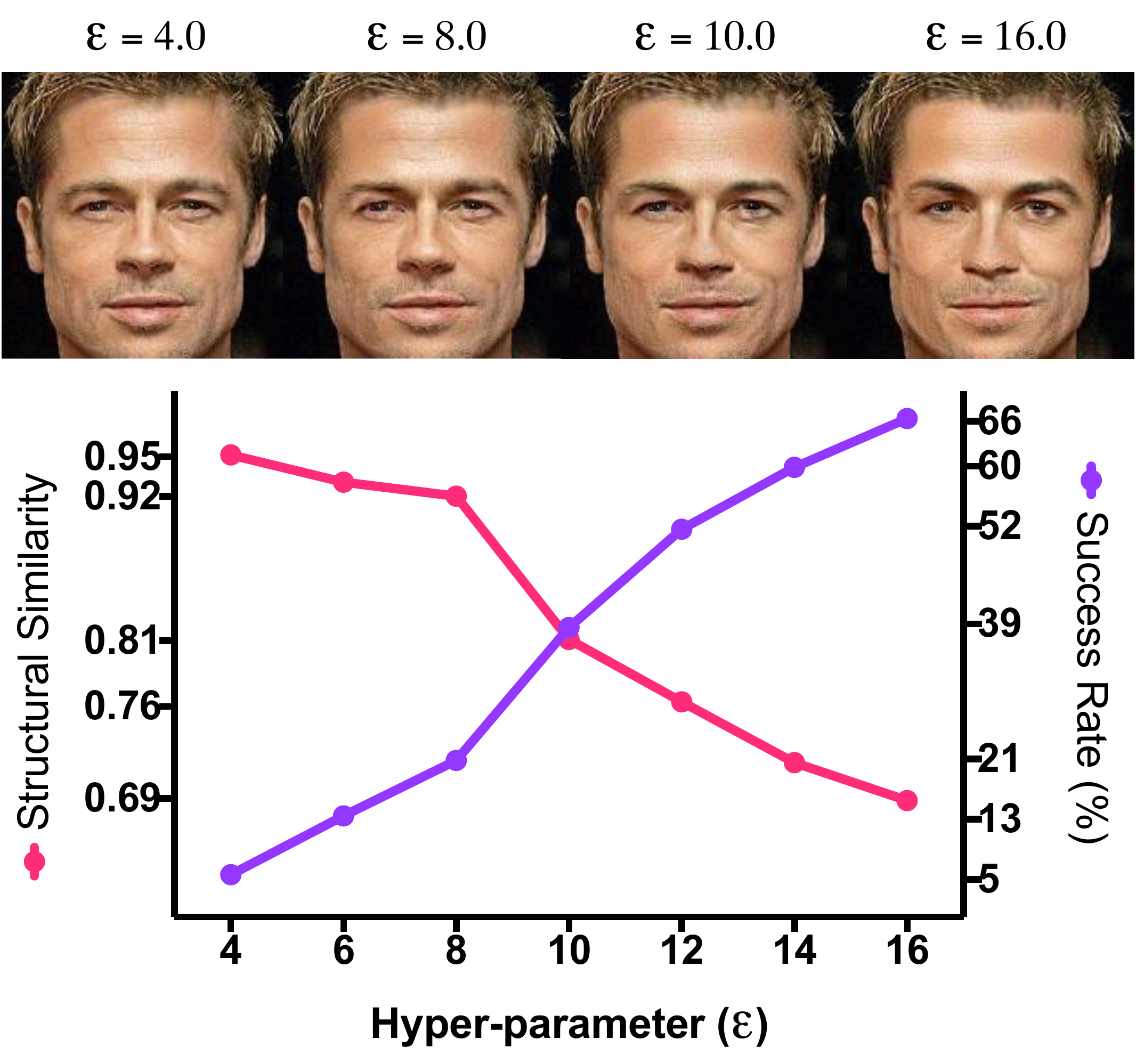}
    \caption{Trade-off between attack success rate and structural similarity for impersonation attacks. We choose $\epsilon = 8.0$.}
    \label{fig:min_eps}
\end{figure}

\vspace{-1em}\paragraph{Experimental Settings}
We use ADAM optimizers in Tensorflow with $\beta_{1} = 0.5$ and $\beta_{2}=0.9$ for the entire network. Each mini-batch consists of $32$ face images. We train AdvFaces for 200,000 steps with a fixed learning rate of $0.0001$. Since our goal is to generate adversarial faces with high success rate, the identity loss is of utmost importance. We empirically set $\lambda_i = 10.0$ and $\lambda_p = 1.0$. We train two separate models and set $\epsilon = 3.0$ and $\epsilon = 8.0$ for obfuscation and impersonation attacks, respectively. All experiments are conducted using Tensorflow r1.12.0 and a NVIDIA Quadro M6000 GPU. Implementations are provided in Appendix~\ref{apdx:implementation}.

\vspace{-1em}\paragraph{Face Recognition Systems} For all our experiments, we employ 5 state-of-the-art face matchers\footnote{All the open-source and COTS matchers achieve 99\% accuracy on LFW under LFW protocol.}. Three of them are publicly available, namely, FaceNet~\cite{facenet}, SphereFace~\cite{sphereface}, and ArcFace~\cite{arcface}. We also report our results on two commercial-off-the-shelf (COTS) face matchers, COTS-A and COTS-B\footnote{Both COTS-A and COTS-B utilize CNNs for face recognition.}. We use FaceNet~\cite{facenet} as the white-box face recognition model, $\mathcal{F}$, during training. \emph{All the testing images in this paper are generated from the same model (trained only with FaceNet) and tested on different matchers.}

\subsection{Comparison with State-of-the-Art} In Table~\ref{tab:recognition}, we find that compared to the state-of-the-art, AdvFaces generates adversarial faces that are similar to the probe. Moreover, the adversarial images attain a high obfuscation attack success rate on 4 state-of-the-art black-box AFR systems in both obfuscation and impersonation settings. AdvFaces learns to perturb the salient regions of the face, unlike PGD~\cite{madry}, FGSM~\cite{goodfellow} which perturbs every pixel in the image. Due to this, we find that, albeit high success rate, the structural similarity between probes and synthesized faces for PGD and FGSM are low. GFLM~\cite{gflm}, on the other hand, geometrically warps the face images and thereby, results in low structural similarity. In addition, the state-of-the-art matchers are robust to such geometric deformation which explains the low success rate of GFLM on face matchers. A$^3$GN is also a GAN-based method, however, fails to achieve a reasonable success rate in an impersonation setting. In Figure~\ref{fig:comparison}, we see that, in addition to high success rate, adversarial faces generated by the proposed method are visually appealing and the differences between probe and synthesized images are hardly distinguishable compared to the baselines.

\subsection{Ablation Study}
In order to analyze the importance of each module in our system, in Figure~\ref{fig:ablation}, we train three variants of AdvFaces for comparison by removing the GAN loss ($\EL_{GAN}$), perturbation loss $\EL_{perturbation}$, and identity loss $\EL_{identity}$, respectively. 
 The GAN loss helps to ensure the visual quality of the synthesized faces are maintained. With the generator alone, undesirable artifacts are introduced. Without the proposed perturbation loss, perturbations in the adversarial mask are unbounded and therefore, leads to a lack in perceptual quality. The identity loss is imperative in ensuring an adversarial image is obtained. Without the identity loss, the synthesized image cannot evade state-of-the-art face matchers. 
We find that every component of AdvFaces is necessary in order to obtain an adversarial face that is not only perceptually realistic but can also evade state-of-the-art face matchers.

\subsection{What is AdvFaces Learning?}
During training, AdvFaces learns to perturb the salient facial regions that can evade the face matcher, $\mathcal{F}$ (FaceNet~\cite{facenet} in our case). This is enforced by $\EL_{perturbation}$ which penalizes large perturbations and thereby, restricts perturbations to only salient pixel locations. In Figure~\ref{fig:visualization}, AdvFaces synthesizes the adversarial masks corresponding to the probes. We then threshold the mask to extract pixels with perturbation magnitudes exceeding $0.40$. It can be inferred that the eyebrows, eyeballs, and nose contain highly discriminative information that an AFR system utilizes to identify an individual. Therefore, perturbing these salient regions are enough to evade state-of-the-art face recognition systems.
\subsection{Effect of Perturbation Amount}
The perturbation hinge loss, $\EL_{perturbation}$ is bounded by a hyper-parameter, $\epsilon$. That is, the $L_2$ norm of the adversarial mask must be at least $\epsilon$. Without this constraint, the adversarial mask becomes an empty image with no changes to the probe. With $\epsilon$, we can observe a trade-off between the attack success rate and the structural similarity between the probe and synthesized adversarial face in Figure~\ref{fig:min_eps}. A higher $\epsilon$ leads to less perturbation restriction. This generates a higher attack success rate at the cost of a lower structural similarity. For an impersonation attack, this implies that the adversarial image may contain facial features from both the hacker and the target. In our experiments, we chose $\epsilon = 8.0$ and $\epsilon = 3.0$ for impersonation and obfuscation attacks, respectively.

\section{Conclusions}
We proposed a new method of adversarial face synthesis,  namely \emph{AdvFaces},  that automatically generates adversarial face images with imperceptible perturbations evading state-of-the-art face matchers. With the help of a GAN, and the proposed perturbation and identity losses, AdvFaces learns the set of pixel locations required by face matchers for identification and only perturbs those salient facial regions (such as eyebrows and nose). Once trained, AdvFaces generates high quality and perceptually realistic adversarial examples that are benign to the human eye but can evade state-of-the-art black-box face matchers, outperforming other state-of-the-art adversarial face methods.

{\small
\bibliographystyle{ieee_fullname}
\bibliography{egbib}
}

\appendix
\section{Implementation Details} 
\emph{AdvFaces} is implemented using Tensorflow r1.12.0. A single NVIDIA Quadro M6000 GPU is used for training and testing.
\label{apdx:implementation}
\paragraph{Data Preprocessing} All face images are passed through MTCNN face detector~\cite{mtcnn} to detect five landmarks (two eyes, nose, and two mouth corners). Via similarity transformation, the face images are aligned. After transformation, the images are resized to $160\times 160$. Before passing into networks, each pixel in the RGB image is normalized by subtracting 127.5 and dividing by 128.
\paragraph{Architecture} Let \texttt{c7s1-k} be a $7\times7$ convolutional layer with $k$ filters and stride $1$. \texttt{dk} denotes a $4\times 4$ convolutional layer with $k$ filters and stride $2$. \texttt{Rk} denotes a residual block that contains two $3\times 3$ convolutional layers. \texttt{uk} denotes a $2\times$
upsampling layer followed by a $5\times 5$ convolutional layer with $k$ filters and stride $1$. We apply Instance Normalization and Batch Normalization to the generator and discriminator, respectively. We use Leaky ReLU with slope 0.2 in the discriminator and ReLU activation in the generator. The architectures of the two modules are as follows:
\begin{itemize}
    \itemsep0em
    \item Generator: \\ 
    \texttt{c7s1-64,d128,d256,R256,R256,R256, u128, u64, c7s1-3}
    \item Discriminator: \\ \texttt{d32,d64,d128,d256,d512}
\end{itemize}
A $1\times 1$ convolutional layer with $3$ filters and stride $1$ is attached to the last convolutional layer of the discriminator for the patch-based GAN loss $\EL_{GAN}$.

\noindent We apply the \texttt{tanh} activation function on the last convolution layer of the generator to ensure that the generated image $\in[-1, 1]$. In the paper, we denoted the output of the tanh layer as an ``adversarial mask'', $\mathcal{G}(x) \in [-1,1]$ and $x \in [-1,1]$. The final adversarial image is computed as $x_{adv} = 2 \times \texttt{clamp}\left[ \mathcal{G}(x) + \left(\frac{x+1}{2}\right) \right]_{0}^{1} -1$. This ensures $\mathcal{G}(x)$ can either add or subtract pixels from $x$ when $\mathcal{G}(x) \neq 0$. When $\mathcal{G}(x)\to0$, then $x_{adv}\to x$.

The overall algorithm describing the training procedure of~\emph{AdvFaces} can be found in Algorithm~\ref{apdx:alg}.

\begin{algorithm}[!ht]
    \caption{Training \emph{AdvFaces} via Adam optimizers. All experiments in this work use $\alpha = 0.0001$, $\beta_1 = 0.5$, $\beta_2 = 0.9$, $\lambda_i = 10.0$, $\lambda_p = 1.0$, $m = 32$, $\epsilon = 3.0$,  and $\epsilon = 8.0$ for obfuscation and impersonation attacks, respectively.}\label{alg:advfaces}
  \begin{algorithmic}[1]
     \Input
     \Desc{$X$}{Training Dataset}
      \Desc{$\mathcal{F}$}{Cosine similarity between an image pair obtained by face matcher}
      \Desc{$\mathcal{G}$}{Generator with weights $\mathcal{G}_\theta$}
      \Desc{$\mathcal{D}$}{Discriminator with weights $\mathcal{D}_\theta$}
      \Desc{$m$}{Batch size}
      \Desc{$\alpha$}{Learning rate}
  \EndInput
    \For{number of training iterations}
        \State \text{Sample a batch of probes $\{x^{(i)}\}_{i=1}^{m} \sim \mathcal{X}$}
        \If{impersonation attack}
            \State Sample a batch of target images ${y^{(i)}} \sim \mathcal{X}$
            \State $\delta^{(i)} = \mathcal{G}((x^{(i)}, y^{(i)})$
        \ElsIf{obfuscation attack}
            \State $\delta^{(i)} = \mathcal{G}(x^{(i)})$
        \EndIf
        \State $x_{adv}^{(i)} = x^{(i)} + \delta^{(i)}$
        \State $\EL_{perturbation} = \frac{1}{m}\left[ \sum_{i=1}^{m}\text{max}\left(\epsilon,||\delta^{(i)}||_2\right)\right]$
        \If{impersonation attack}
            \State $\EL_{identity} = \frac{1}{m}\left[\sum_{i=1}^{m}\mathcal{F}\left(x^{(i)}, x_{adv}^{(i)}\right)\right]$
        \ElsIf{obfuscation attack}
            \State $\EL_{identity} = \frac{1}{m}\left[\sum_{i=1}^{m}\left(1 - \mathcal{F}\left(y^{(i)}, x_{adv}^{(i)}\right)\right)\right]$
        \EndIf
        \State $\EL^{\mathcal{G}}_{GAN} = \frac{1}{m}\left[\sum_{i=1}^{m}log\left(1-\mathcal{D}(x_{adv}^{(i)}) \right)\right]$
        \State $\EL^{\mathcal{D}} = \frac{1}{m}\left[\sum_{i=1}^{m}log\left(\mathcal{D}(x^{(i)}) \right) +  log\left(1-\mathcal{D}(x_{adv}^{(i)}) \right)\right]$
        \State $\EL^{\mathcal{G}} = \EL^{\mathcal{G}}_{GAN} + \lambda_i \EL_{identity} + \lambda_p \EL_{perturbation}$
        \State $\mathcal{G}_{\theta} = \text{Adam}(\triangledown_{\mathcal{G}}\EL^{\mathcal{G}}, \mathcal{G}_{\theta},\beta_{1},\beta_{2})$
        \State $\mathcal{D}_\theta = \text{Adam}(\triangledown_{\mathcal{D}}\EL^{\mathcal{D}}, \mathcal{D}_{\theta},\beta_{1},\beta_{2})$
      \EndFor
    \end{algorithmic}
    \label{apdx:alg}
\end{algorithm}
\section{Structural Similarity} Image comparison techniques, such Mean Squared Error (MSE) or Peak Signal-to-Noise Ratio (PSNR), estimate the absolute errors, disregarding the \emph{perceptual} differences; on the other hand, SSIM is a perception-based model that considers image differences as perceived change in structural information, while also incorporating important perceptual phenomena, including both luminance masking and contrast masking terms. For instance, consider the image pair comprising of two images of Ming Xi. We can notice that perceptually, the image pairs are similar, but this perceptual similarity is not reflected appropriately in MSE and PSNR. Since, SSIM is a normalized similarity metric, it is better suited for our application where a face image pair is subjectively judged by human operators.
\begin{figure}[!h]
    \centering
    \subfloat[Probe]{\includegraphics[width=0.49\linewidth]{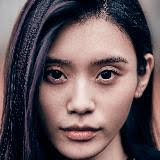}}\hfill
    \subfloat[Adversarial]{\includegraphics[width=0.49\linewidth]{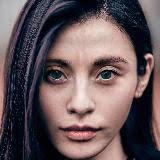}}
    \caption*{SSIM: 00.96 \quad\quad MSE: 40.82 \quad\quad PSNR: 32.02}
    \label{fig:mse_ssim}
\end{figure}
\section{Effect on Cosine Similarity}
\begin{figure}[!h]
    \centering
    \subfloat{\includegraphics[width=\linewidth]{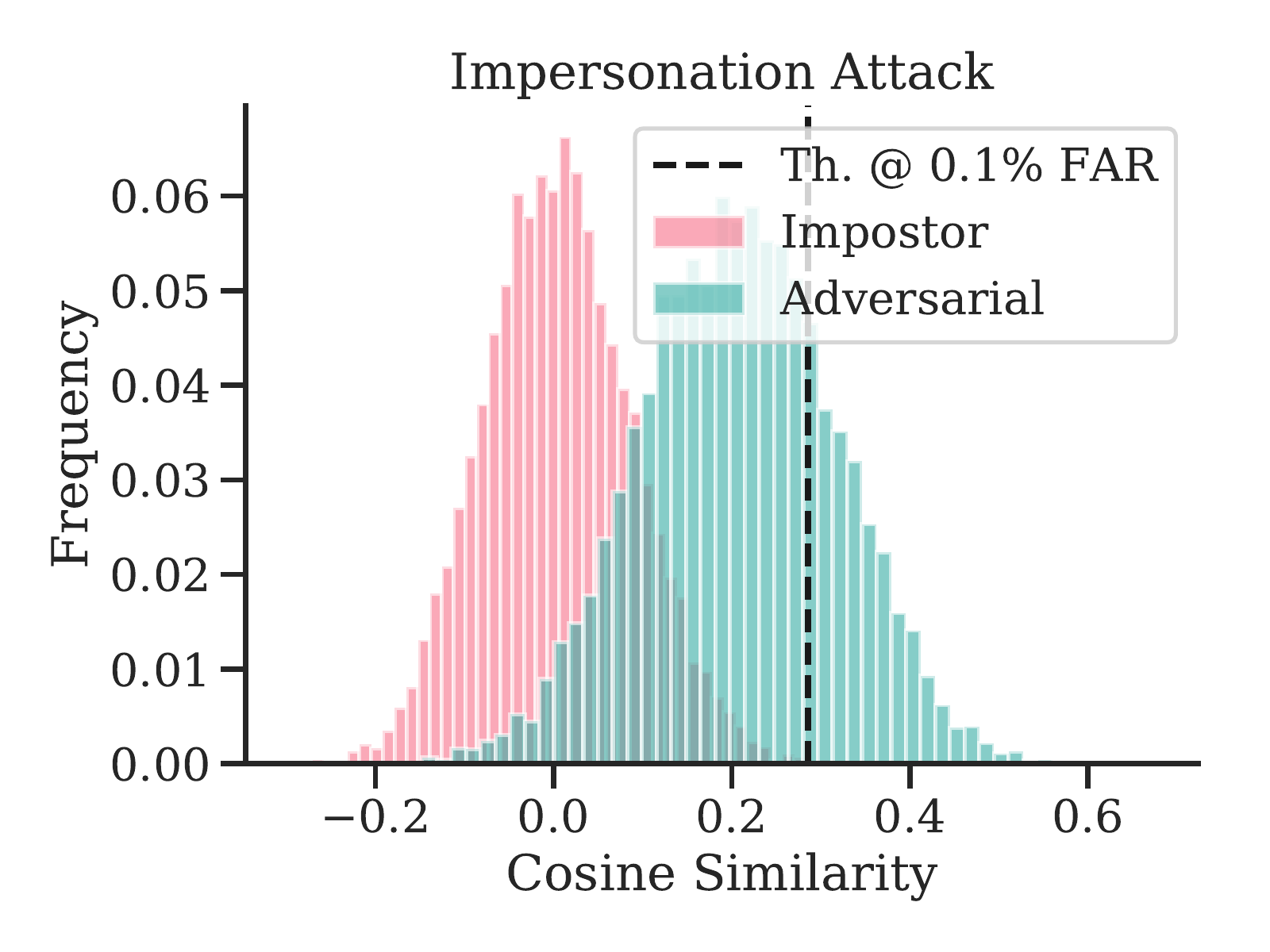}}\\[-1em]
    \subfloat{\includegraphics[width=\linewidth]{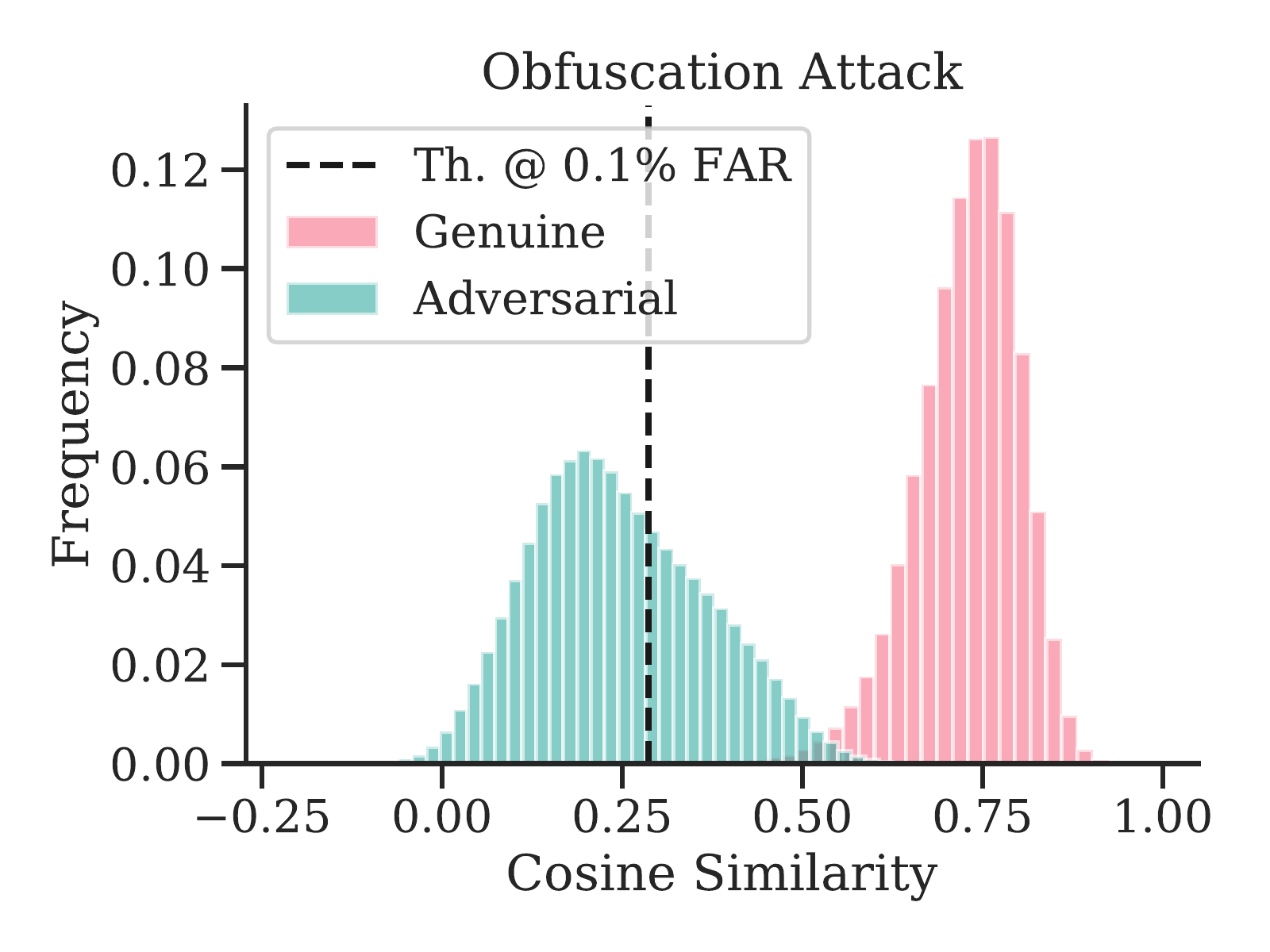}}
    \caption{Shift in cosine similarity scores for ArcFace~\cite{arcface} before and after adversarial attacks generated via AdvFaces.}
    \label{fig:sim_scores}
\end{figure}

In Figure~\ref{fig:sim_scores}, we see the effect on cosine similarity scores when adversarial face images synthesized by AdvFaces is introduced to a black-box face matcher, ArcFace~\cite{arcface}. A majority ($64.53\%$) of the scores fall below the threshold at $0.1\%$ FAR causing the AFR system to falsely reject under obfuscation attack. In the impersonation attack setting, the system falsely accepts $24.30\%$ of the image pairs.

\section{Baseline Implementation Details}
All the state-of-the-art baselines in the paper are implementations proposed specifically for evading face recognition systems. 

\paragraph{FGSM~\cite{goodfellow}} We use the Cleverhans implementation\footnote{\url{https://github.com/tensorflow/cleverhans/tree/master/examples/facenet_adversarial_faces}} of FGSM on FaceNet. This implementation supports both obfuscation and impersonation attacks. The only modification was changing $\epsilon = 0.01$ to $\epsilon = 0.08$ in order to create more effective attacks.
\paragraph{PGD~\cite{madry}} We use a variant of PGD proposed specifically for face recognition systems~\footnote{\url{https://github.com/ppwwyyxx/Adversarial-Face-Attack}}. Originally, this implementation is proposed for impersonation attacks, however, for obfuscation we randomly choose a target other than genuine subject. We do not make any modifications to the parameters.
\paragraph{GFLM~\cite{gflm}} Code for this landmark-based attack synthesis method is publicly available~\footnote{\url{https://github.com/alldbi/FLM}}. This method relies on softmax probalities implying that the training and testing identities are fixed. Originally, the classifier is trained on CASIA-WebFace. However, for a fairer evaluation, we trained a face classifier on LFW and then ran the attack.
\paragraph{A$^3$GN~\cite{song}} To the best of our knowledge, there is no publicly available implementation of A$^3$GN. Our implementation is included with our open-source code~\footnote{[Link omitted for blind review]}. We made the following modifications to achieve an effective baseline:
\begin{itemize}\itemsep0em
    \item The authors originally used ArcFace~\cite{arcface} as the target model. Since all other baselines employ FaceNet as the target model, we also used FaceNet for training A$^3$GN.
    \item Originally, a cycle-consistency loss was proposed for content preservation. However, we were not able to reproduce this and therefore, opted for the same $L_1$ norm loss, but without the second generator. This greatly helps in the visual quality of the generated adversarial image. That is, we modified Equation 3~\cite{song}, from $\EL_{rec} = \IE_{x,z}\left[||x-G_2(G_1(x,z))||_1\right]$ to $\EL_{rec} = \IE_{x,z}\left[||x-G_1(x,z)||_1\right]$
\end{itemize}

\end{document}